\documentclass[letterpaper, 10 pt, conference]{ieeeconf}  

\IEEEoverridecommandlockouts                              

\usepackage{graphics} 
\usepackage{epsfig} 
\usepackage{mathptmx} 
\usepackage{times} 
\usepackage{amsmath} 
\usepackage{amssymb}  
\usepackage{booktabs}
\usepackage{multirow}

\usepackage[utf8]{inputenc}
\usepackage{kotex} 
\usepackage{graphicx}
\usepackage{bm}
\usepackage{amsfonts}
\usepackage{hyperref}
\usepackage{stfloats} 
\usepackage{caption}
\usepackage{subcaption}
\usepackage{adjustbox} 

\title{\LARGE \bf Leveraging Text-Driven Semantic Variation\\for Robust OOD Segmentation}

\author{Seungheon Song$^{1}$ and Jaekoo Lee$^{1}$$^{*}$%
\thanks{$^{1}$Seungheon Song and Jaekoo Lee are with the College of Computer Science, Kookmin University, Seoul, Republic of Korea}%
\thanks{$^{*}$Corresponding author: Jaekoo Lee (jaekoo@kookmin.ac.kr)}}%

\begin{document}

\maketitle
\thispagestyle{empty}
\pagestyle{empty}


\begin{figure*}[b]
    \centering
    \includegraphics[width=\textwidth]{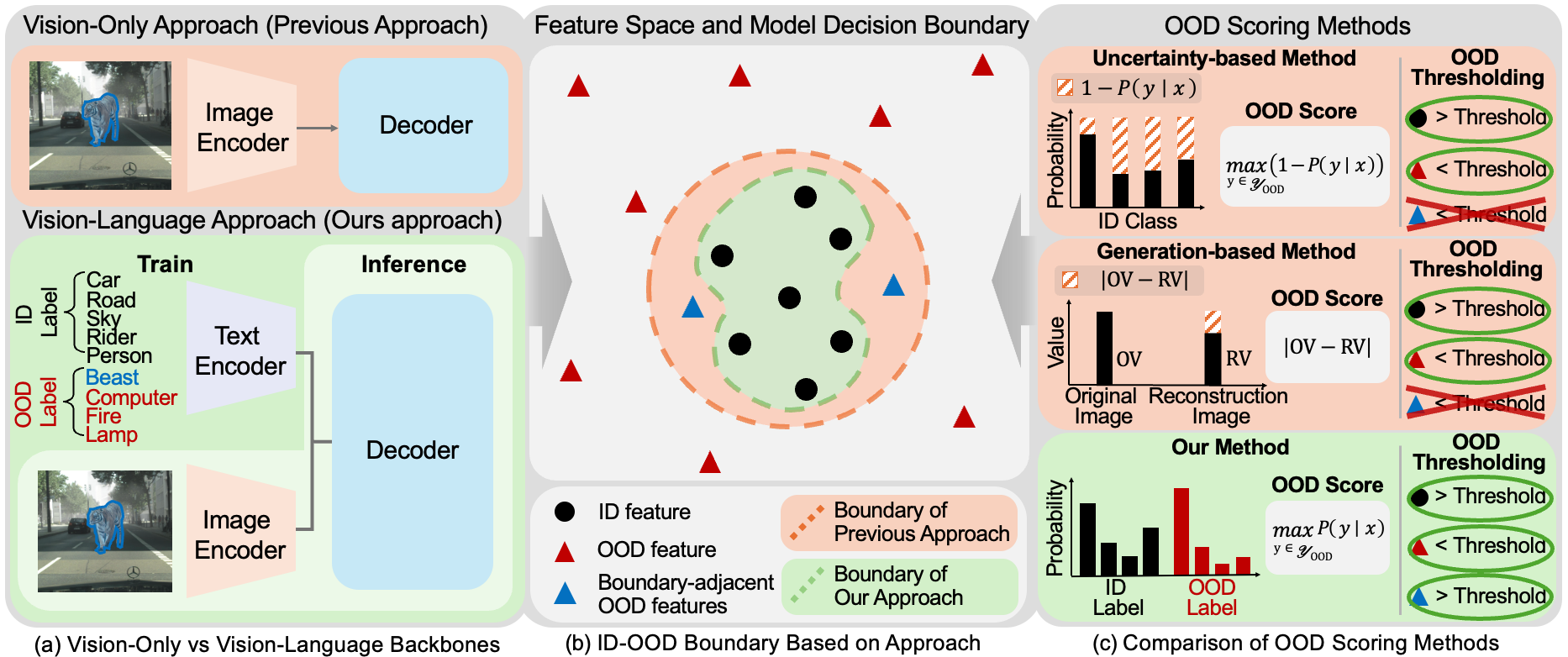} 
    \caption{Overview of limitations in existing OOD segmentation approaches and advantages of the proposed approach. (a) Existing methods often rely solely on visual information (vision-only), whereas our vision-language approach incorporates textual cues in addition to images. (b) By leveraging semantic information from text, our method learns clearer decision boundaries in the joint ID and OOD feature space. (c) Unlike uncertainty- or generation-based methods that use only visual cues, our approach leverages textual knowledge to achieve more reliable OOD scoring.}

    \label{fig:teaser}
\end{figure*}

\begin{abstract}

In autonomous driving and robotics, ensuring road safety and reliable decision-making critically depends on out-of-distribution (OOD) segmentation. While numerous methods have been proposed to detect anomalous objects on the road, leveraging the vision-language space--which provides rich linguistic knowledge--remains an underexplored field. We hypothesize that incorporating these linguistic cues can be especially beneficial in the complex contexts found in real-world autonomous driving scenarios.

To this end, we present a novel approach that trains a \emph{Text-Driven OOD Segmentation} model to learn a semantically diverse set of objects in the vision-language space. Concretely, our approach combines a vision-language model's encoder with a transformer decoder, employs \emph{Distance-Based OOD prompts} located at varying semantic distances from in-distribution (ID) classes, and utilizes \emph{OOD Semantic Augmentation} for OOD representations. By aligning visual and textual information, our approach effectively generalizes to unseen objects and provides robust OOD segmentation in diverse driving environments.

We conduct extensive experiments on publicly available OOD segmentation datasets such as Fishyscapes, Segment-Me-If-You-Can, and Road Anomaly datasets, demonstrating that our approach achieves state-of-the-art performance across both pixel-level and object-level evaluations. This result underscores the potential of vision-language–based OOD segmentation to bolster the safety and reliability of future autonomous driving systems.

\end{abstract}


\section{INTRODUCTION}

Semantic segmentation is essential for autonomous driving because it classifies environment elements (e.g., roads, pedestrians, vehicles) at the pixel level. However, out-of-distribution (OOD) objects—unseen during training—frequently appear in real-world driving scenarios. Existing segmentation models often misclassify these OOD objects as in-distribution (ID), increasing the risk of collisions. Previous OOD segmentation methods~\cite{PEBAL,densehybrid,synboost, S2M, RPL, Mask2Anomaly} typically rely on vision-only approach (see the top of Figure\ref{fig:teaser}(a)) or specialized OOD datasets, but they still struggle to generalize effectively (see Figure~\ref{fig:teaser}(b)).

To address these issues, we propose: i) \emph{Text-Driven OOD Segmentation}: By integrating CLIP’s vision-language modeling~\cite{clip} into Mask2Former~\cite{mask2former}, our method processes both textual and visual cues for more robust OOD recognition. ii) \emph{Distance-Based OOD Prompts}: Using WordNet~\cite{WordNet}, we generate multiple OOD queries based on their semantic distance from ID classes, improving OOD segmentation accuracy in the Mask2Former Transformer decoder. iii) \emph{OOD Semantic Augmentation}: Instead of inserting external objects (e.g., from COCO ~\cite{coco}), we apply self-attention–based feature adjustments to diversify OOD representations, enabling better handling of unseen anomalies in real-world driving.

By combining a text-driven OOD segmentation, distance-based OOD prompt, and OOD semantic augmentation, our approach delivers a robust solution for OOD segmentation in autonomous driving. Extensive evaluations on multiple datasets confirm its superior performance and generalization compared to existing methods. Overall, our text-driven strategy significantly enhances OOD segmentation robustness and marks a promising step toward vision-language–based perception in autonomous driving.

\section{RELATED WORK}

\subsection{Semantic Segmentation}
\noindent\textbf{Vision-Only Segmentation.}
For autonomous driving scenarios, pixel-level segmentation has predominantly relied on vision-only methods based on CNNs~\cite{ResNet,DeepLab} and Vision Transformers (ViTs)~\cite{ViT, swin, ha2024leveraging, kwon2023mobile}. These models achieve high performance by learning visual patterns from large-scale image datasets paired with segmentation labels. However, they depend heavily on predefined categories, making it challenging to accurately recognize OOD objects that frequently arise in real-world driving.

\noindent\textbf{Vision-Language Models (VLMs) for Segmentation.}
Recently, the advent of large-scale VLMs trained on image-text pairs, such as CLIP\cite{clip} and ALIGN~\cite{ALIGN}, has spurred active research on leveraging textual cues for segmentation~\cite{DenseCLIP,maskclip,SAM-CLIP,FC-CLIP,VLTSeg}. Unlike vision-only approaches, VLM-based methods can incorporate richer semantic information and flexibly handle OOD data~\cite{ghosh2024exploringfrontiervisionlanguagemodels, zhang2024visionlanguagemodelsvisiontasks}. 


\noindent\textbf{Prompt Tuning for VLMs.}
Despite the potential of applying VLMs to downstream tasks, designing text prompts has emerged as a critical challenge. Early approaches relied on handcrafted prompts, but these often failed to capture the diverse environmental contexts and fine-grained object details encountered in autonomous driving. To address this issue, prompt tuning techniques~\cite{coop, cocoop} have been proposed to automatically adapt pretrained VLM knowledge to specific tasks. In the segmentation domain, several studies explore soft prompts~\cite{TQDM, MTA-CLIP, CLIPSeg, SSPrompt, lee2025controllable} to learn detailed features that are difficult to capture via hard prompts, thereby improving adaptability to various environmental changes and OOD objects.

While vision-only methods excel at learning rich visual representations, they face clear limitations in handling OOD scenarios. In contrast, VLM-based segmentation offers enhanced flexibility and generality by incorporating textual cues, and prompt tuning further eases adaptation to diverse settings. Building on this momentum, we propose a novel approach that effectively addresses the challenge of OOD segmentation in autonomous driving environments.

\subsection{Robust OOD in Autonomous Driving Scenarios}
Autonomous driving systems must handle unexpected objects and conditions that do not appear in the training data. Conventional segmentation methods focus primarily on known classes (e.g., pedestrians, vehicles) but often fail to detect new or anomalous objects \cite{robotood, li2020outofdistributiondetectionskinlesion}. Recent approaches for OOD handling in driving scenarios include:

\noindent\textbf{Uncertainty-Based Methods} \cite{hendrycks2018baselinedetectingmisclassifiedoutofdistribution, tian2021weaklysupervisedvideoanomalydetection, synboost, liang2020enhancingreliabilityoutofdistributionimage, lee2018trainingconfidencecalibratedclassifiersdetecting}: These estimate pixel-level uncertainty (see Fig.\ref{fig:teaser}(c)) but often struggle to delineate object boundaries accurately. 

\noindent\textbf{Reconstruction-Based Methods} \cite{PEBAL, RoadAnomaly, haldimann2019iimaginederrordetection, xia2020synthesizecomparedetectingfailures}: These compare the input image with a reconstructed version to detect anomalies (see Fig.\ref{fig:teaser}(c)), but their effectiveness heavily depends on generation quality. 

\noindent\textbf{Attention-Based Methods} \cite{RbA, Mask2Anomaly, mask2former, SAM, S2M}: These utilize transformer decoders to segment objects rather than individual pixels. For instance, \cite{Mask2Anomaly, RbA} fuse class and mask predictions, while \cite{S2M} integrates a SAM decoder to enhance object-level OOD segmentation.

\noindent\textbf{Augmentation-Based Methods} \cite{PEBAL, Mask2Anomaly,RPL, POC}: In the context of OOD segmentation for autonomous driving, outlier exposure methods introduce OOD samples (often from auxiliary datasets) into ID scenes~\cite{PEBAL, Mask2Anomaly,RPL}. While this can improve OOD detection, it may also introduce unrealistic OOD examples, cause excessive domain shifts, or limit OOD diversity. Recent studies employ generative models (e.g., Stable Diffusion~\cite{StableDiffusion}) to synthesize more realistic OOD data~\cite{POC}, but such approaches often entail high computational cost.

Despite these advances, most existing OOD methods focus solely on visual cues. To overcome this limitation, we propose coupling the CLIP~\cite{clip} with a attention-based segmentation model~\cite{mask2former}, enhanced by distance-aware prompt tuning to better differentiate between ID and OOD classes.

\section{METHOD}

\begin{figure*}[t]
    \centering
    \includegraphics[width=\textwidth]{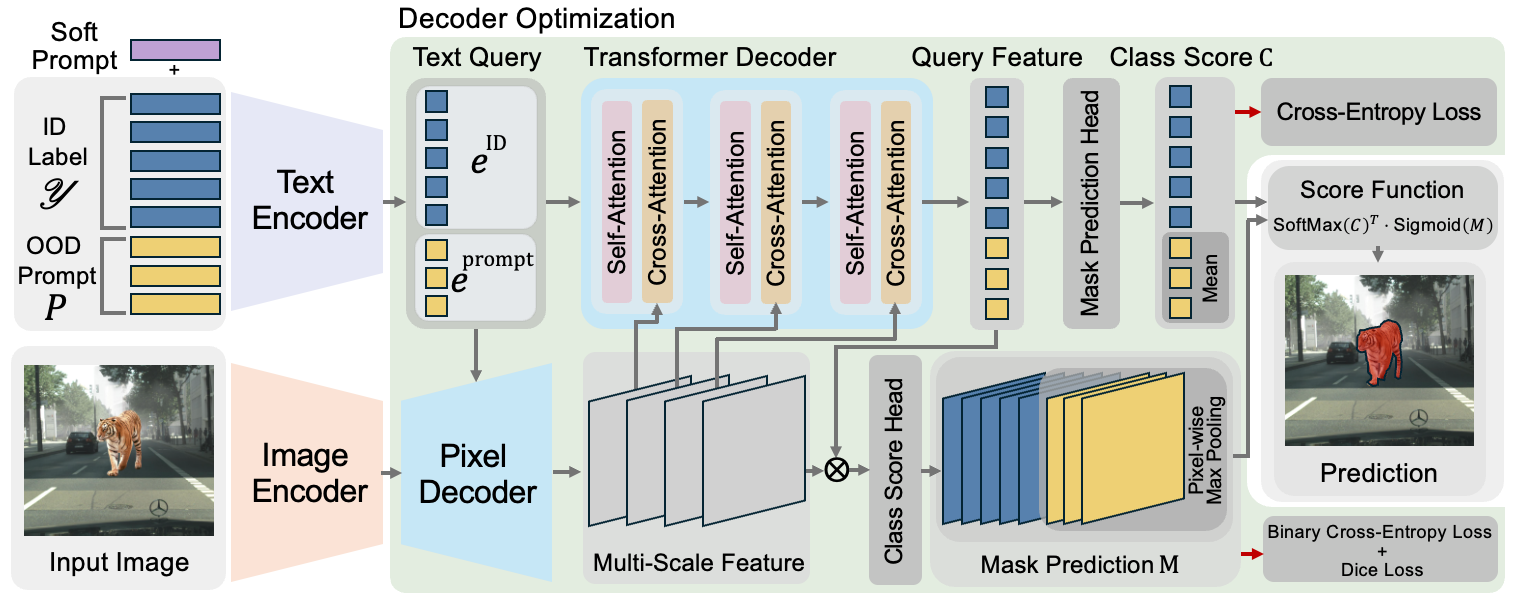} 
    \caption{An overview architecture of the proposed method}
    \label{fig:overview}
\end{figure*}


The objective of OOD segmentation is to accurately detect and segment regions or objects belonging to categories not seen during training. Formally, given an input image $\mathbf{X} \in \mathbb{R}^{H \times W \times C}$ the model produces two outputs for each pixel $(i, j)$: a predicted class $pred_{(i,j)}$ and an OOD confidence score $s_{(i,j)}$.

Let $\mathcal{Y}_{ID}$ be the set of ID labels and $\mathcal{Y}_{OOD}$ be the set of OOD labels. The pixel-level classification function $f_{OOD}$ is defined as:
\begin{equation}
    f_{OOD}: \mathbf{X} \to \mathbf{Y}, \quad 
    \mathbf{Y} = \{ pred_{(i,j)} \in \mathcal{Y}_{ID} \cup \mathcal{Y}_{OOD} \}
\end{equation}
\noindent where each pixel $(i, j)$ is predicted to be either ID or OOD.
Subsequently, a score function $f_{score}$ maps the output $Y$ to a set of pixel-wise confidence scores: 
\begin{equation}
    f_{score}: \mathbf{Y} \to \mathbf{S}, \quad 
    \mathbf{S} = \{ s_{(i,j)} \in [0,1] \}
\end{equation}
\noindent where $s_{(i,j)}$ indicates the probability that pixel $(i, j)$ is OOD. To achieve reliable OOD segmentation, the model must not only distinguish OOD objects from the existing label space but also assign high confidence scores to OOD pixels.

We present a novel method that leverages vision-language modeling to enhance OOD segmentation performance. By combining \emph{Text-Based OOD Segmentation}, \emph{Distance-Based OOD Prompts}, and \emph{OOD Semantic Augmentation}, our method surpasses existing alternatives in performance and efficiency for detecting and segmenting OOD objects.




\begin{figure}[t]
    \centering
    \includegraphics[width=\columnwidth]{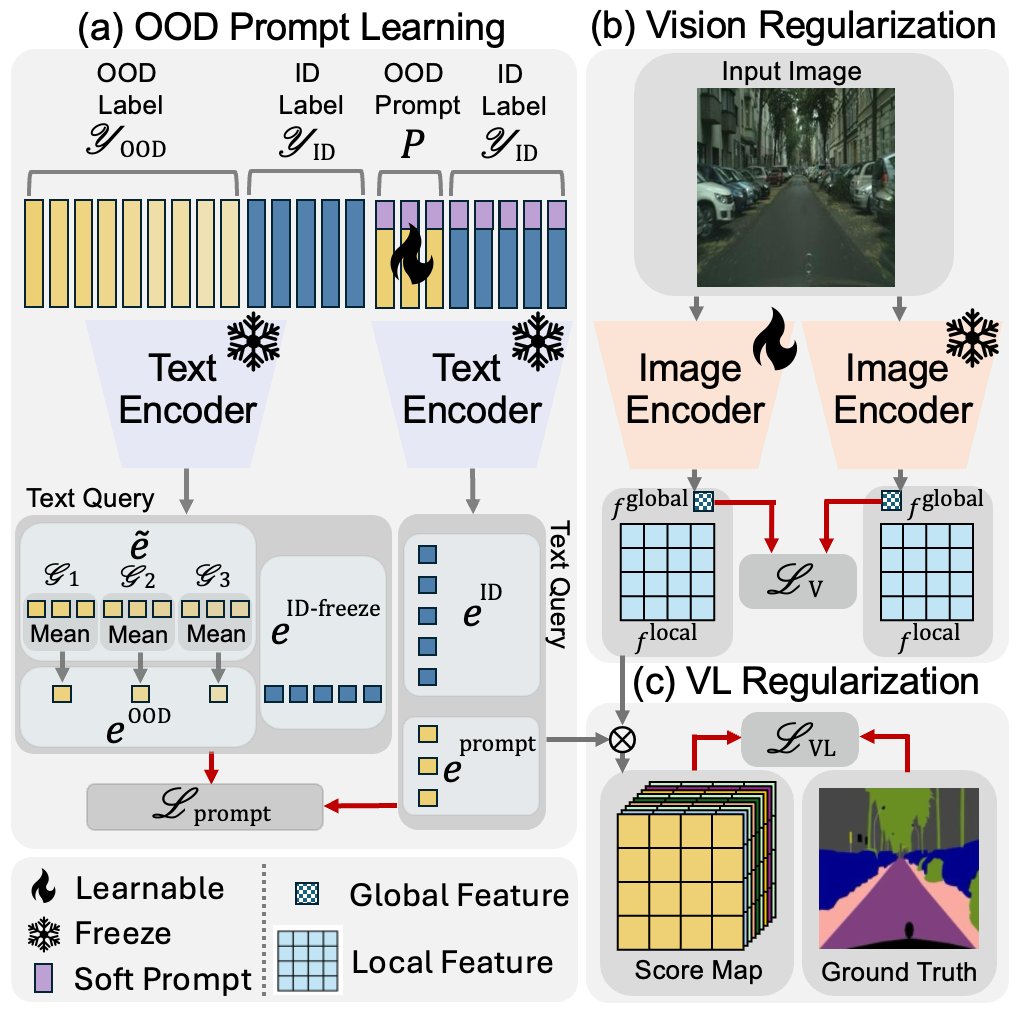} 
    \caption{Overview of strategies for improving \emph{Text-Driven OOD Segmentation}: (a) \emph{Distance-Based OOD Prompts}: Learns OOD prompts placed at various semantic distances from each ID label, thereby enhancing the model’s ability to handle diverse unknown categories. (b) \emph{Vision Regularization}: Preserves the pretrained knowledge of the image encoder by minimizing deviations from its original vision-language alignment. (c) \emph{Vision-Language Regularization}: Extends pixel-level vision-language knowledge in the VLM, enabling more comprehensive semantic understanding for improved OOD detection.}
    \label{fig:regularization}
\end{figure}





\subsection{Text-driven OOD Segmentation} 
Figure~\ref{fig:overview} illustrates our \emph{Text-Driven OOD Segmentation} architecture. We integrate the vision encoder and text encoder from the CLIP~\cite{clip}, pretrained on extensive image–text pairs, into the Mask2Former~\cite{mask2former}'s pixel decoder and a transformer decoder. A \emph{backbone regularization} strategy preserves the pretrained vision–language alignment, while our \emph{decoder optimization} strategy refines pixel-level OOD boundaries. This two-pronged architecture maintains the expressive semantic features acquired during pretraining while adapting the segmentation head to the specialized demands of OOD segmentation.

By ensuring consistency between the text and visual branches, our method naturally accommodates both ID and OOD concepts. The text encoder transforms OOD labels into query embeddings that guide the transformer decoder in isolating anomalous objects at the mask level. This approach minimizes interference in the pretrained embedding space, systematically extending the model’s capacity to detect, localize, and segment OOD instances.

This ensures that the trained encoder remains close to the original vision-language–aligned representations while adapting to new data.





Our \emph{backbone regularization} strategy includes two loss functions, vision regularization $\mathcal{L}_{\text{V}}$ and vision-language regularization $\mathcal{L}_{\text{VL}}$. $\mathcal{L}_{\text{V}}$ (Fig.\ref{fig:regularization}(b)) is defined as follows:

\begin{equation}
    \mathcal{L}_{\text{V}} = \| f^{\text{global-freeze}} - f^{\text{global}} \|_2^2
\end{equation}

\noindent applies an $L_2$ norm between the global feature $f_{\text{global}}$ from the trainable image encoder and the global feature $f^{\text{global-freeze}}$ from the frozen image encoder of CLIP\cite{clip}. 

$\mathcal{L}_{\text{VL}}$ (Fig.~\ref{fig:regularization}(c)) is also defined as follows:


\begin{equation}
    \mathcal{L}_{\text{VL}} = \text{CrossEntropyLoss}(\text{Softmax}(S_{i,j,k}), Y_{i,j,k})
\end{equation}

\noindent where $ S_{i,j,k} = f^{\text{local}}_{i,j} \cdot e_k$ is the score map, representing the similarity between the local feature $f^{\text{local}}_{i,j}$ extracted from the image encoder at pixel $(i,j)$ and the text embedding $e_k$ . Here, $e_k$ denotes the text embedding of class $k$. $Y_{i,j,k}$ represents the ground-truth at pixel $(i,j)$ as defined as follows:
\begin{equation}
    Y_{i,j,k} =
    \begin{cases} 
        k, & \text{if } k \leq |\mathcal{Y}_{ID}| \\[3pt]
        K+1, & \text{if } k > |\mathcal{Y}_{ID}|
    \end{cases}
\end{equation}

\noindent where $K$ is the number of ID labels.

Thus, the model preserves rich language knowledge crucial for distinguishing OOD objects.

Figure~\ref{fig:overview} illustrates how text embeddings serve as text queries for both the pixel decoder and the transformer decoder, bridging visual and textual information. As our \emph{decoder optimization} strategy, the pixel decoder first processes image embeddings through self-attention to form multi-scale features, then performs cross-attention with the text queries to incorporate semantic cues. Simultaneously, the transformer decoder applies self-attention and cross-attention, using text queries to extract object-specific features across multi-scale features. The resulting query features feed into a Class Score Head (optimized via cross-entropy) and a Mask Prediction Head (optimized via binary cross-entropy plus Dice loss). 

Consequently, the model refines object boundaries and effectively separates OOD pixels from ID classes in driving environments, harnessing both visual and textual cues for robust segmentation.


\begin{figure}[t]
    \centering
    \includegraphics[width=\columnwidth]{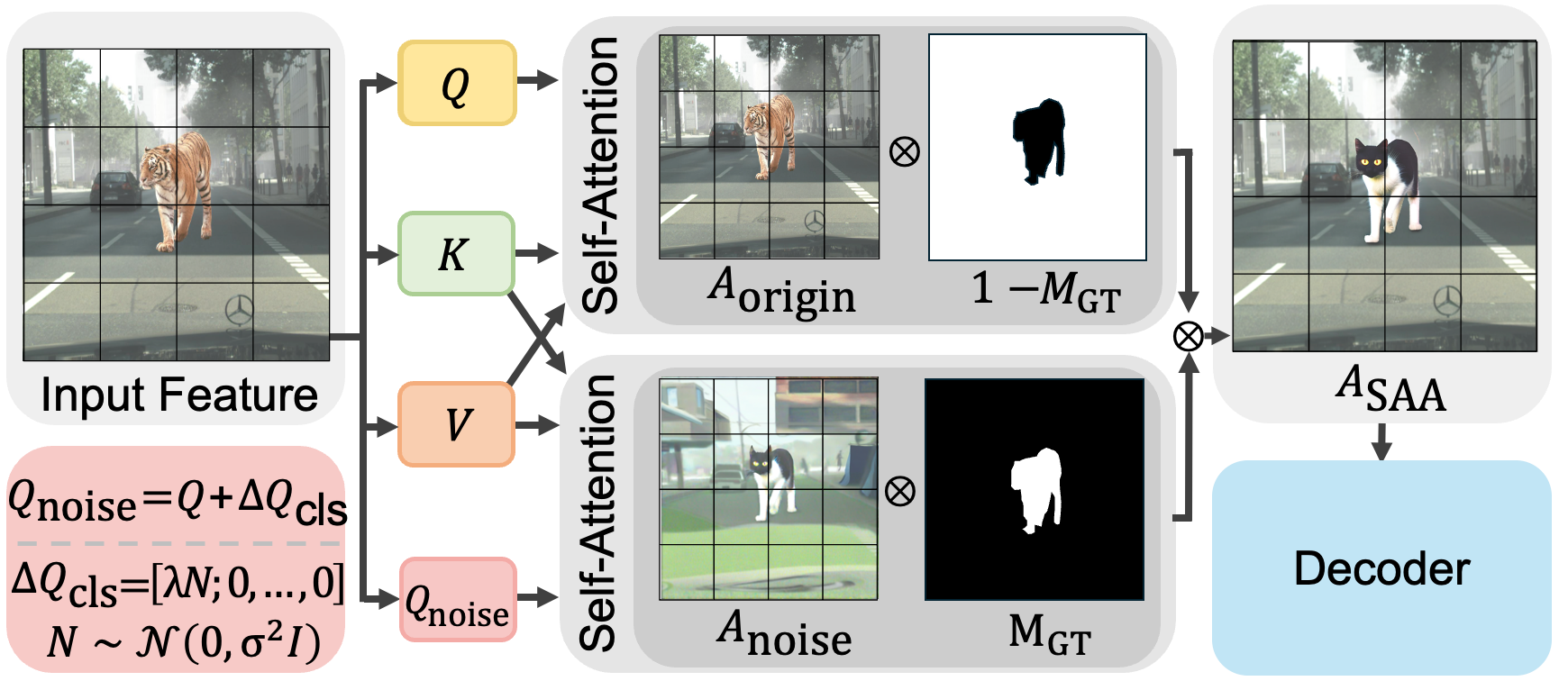} 
    \caption{An overview of our proposed \emph{Semantically Augmented Attention} $A_{\text{SAA}}$ mechanism}
    \label{fig:SAA}
\end{figure}

\subsection{Distance-Based OOD Prompts} 






Previous OOD segmentation approaches rely solely on visual information, limiting their access to diverse OOD representations and often resulting in suboptimal decision boundaries (Figure~\ref{fig:teaser}(b)). To overcome these shortcomings, we introduce a \emph{Distance-Based OOD Prompts} for VLM, which leverages semantically distant OOD labels--span various distances from ID labels in the vision-language embedding space--to enhance pixel-level segmentation of unknown objects. The process proceeds in the following two steps:





\subsubsection{Generating OOD Labels}

First, we employ the NegMining algorithm~\cite{negativelabelguidedood} on the large-scale WordNet corpus~\cite{WordNet}, extracting an OOD label set $\mathcal{Y}_{OOD}$ that lies sufficiently far from the ID label set $\mathcal{Y}_{ID}$. For each ID label $y_i$ in $\mathcal{Y}_{ID} = \{y_1, y_2, ..., y_I\}$, we measure the semantic distance to candidate words $\tilde{y}_j$ sampled from a corpus $\mathcal{Y}_{corpus} = \{\tilde{y}_1, \tilde{y}_2, ..., \tilde{y}_J\}$. Both the ID label and the candidate word are embedded via CLIP’s text encoder $T(\cdot)$:

\begin{equation}
    e^{ID}_i = T(y_i), \quad 
    \tilde{e}_j = T(\tilde{y}_j)
\end{equation}
\noindent where $e^{ID}_i$ denotes the embedding with a soft prompt, and $e^{ID-freeze}_i$ denotes the embedding without one. We compute cosine similarities for randomly sampled words in the corpus, then select OOD label candidates by filtering those in the bottom 5\% of similarity. This strategy excludes extreme outliers while ensuring stable negative label selection.
Finally, we choose the top $M$ candidates from the filtered set to form the OOD label set OOD $\mathcal{Y}_{OOD}$. These labels maintain sufficient semantic distance from the ID labels and serve as the foundation for \emph{Learning Distance-Based OOD Prompts} with broader coverage of unknown objects.



\subsubsection{Learning Distance-Based OOD Prompts}
As shown in Figure~\ref{fig:regularization}(a), we split the OOD labels into $N$ groups according to their semantic distances from the ID labels. Each group $\mathcal{G}_n$ is determined by a range of distances from the ID labels, and we average the text embeddings of all OOD labels in $\mathcal{G}_n$ to obtain $e_n^{\text{OOD}}$ as follows:


\begin{equation}
    e_n^{\text{OOD}} = \frac{1}{|\mathcal{G}_n|} \sum_{\tilde{y}_j \in \mathcal{G}_n} \tilde{e}.
\end{equation}
Next, we initialize a learnable prompt vector $P_n$ and feed it into CLIP’s text encoder~\cite{clip} to generate $e_n^{\text{prompt}}$ as follows:
\begin{equation}
    e_n^{\text{prompt}} = T(P_n).
\end{equation}

In learning stage, we concatenate $e^{\text{OOD}}$ and $e^{\text{ID-freeze}}$, then compute the similarity between the \emph{concat}\{$e^{\text{OOD}}$, $e^{\text{ID-freeze}}$\} and $e_n^{\text{ID}}$. Here, we assign a ground-truth label of 1 if they share the same $i$, or 0 otherwise, such that the model learns to increase similarity for identical $i$ and decrease it for different $i$. Likewise, we apply $e^{\text{prompt}}$ in a similar manner for each $n$. Using cross entropy loss, the OOD prompts can effectively capture a broad spectrum of distance relationships to ID labels—relationships that are otherwise difficult to learn using only hard prompts.


\subsection{OOD Semantic Augmentation}
Effective data augmentation is essential for improving OOD object segmentation. Prior approaches typically rely on either (1) inserting OOD objects from auxiliary datasets, which is computationally cheap but offers limited diversity, or (2) synthesizing objects via generative models, which provides higher diversity at the cost of increased computation.

To address these limitations, we propose \emph{Semantically Augmented Attention} $A_{\text{SAA}}$ (Figure~\ref{fig:SAA}(b)), which selectively modifies the semantic features of OOD regions using an extended self-attention mechanism while preserving the features of ID objects.
Specifically, we compute two parallel attentions--$A_{\text{noise}}$, which injects noise into OOD objects, and $A_{\text{origin}}$, the standard self-attention output--to enable the model to generalize more effectively to a broader range of OOD objects~\cite{kim2024instance, kim2023bridging}.

First, to semantically modify OOD objects, we define \( A_{\text{noise}} \) such that the CLS token’s global information is perturbed. Specifically, we add a normally distributed noise \(\Delta Q_{\text{cls}}\) to the CLS token feature (the 0th index in the token sequence) to create a new query matrix \(Q_{\text{noise}}\):
\begin{equation}
    Q_{\text{noise}} \;=\; Q \;+\;\Delta Q_{\text{cls}}, 
    \quad
    \Delta Q_{\text{cls}} \;=\; [\,\lambda N;\; 0, \dots, 0],
\end{equation}
where \(\lambda\) is a hyperparameter that controls the noise intensity, and $N \;\sim\; \mathcal{N}(0, \sigma^2 I)$ represents the normally distributed noise for semantic transformation. We then compute the \(A_{\text{noise}}\) as:
\begin{equation}
    A_{\text{noise}} \;=\; \text{Softmax}\!\Bigl( \tfrac{Q_{\text{noise}} K^T}{\sqrt{d}} \Bigr) V.
\end{equation}

Consequently, as illustrated in Figure~\ref{fig:SAA}(b), we apply \(A_{\text{origin}}\) to ID object regions and \(A_{\text{noise}}\) to OOD object regions, thereby selectively injecting noise into only the OOD regions. Formally, \(A_{\text{SAA}}\) is defined as follows:

\begin{equation}
    A_{\text{SAA}} =
    \begin{cases}
        M_{\text{GT}} \odot A_{\text{noise}} 
        \;+\;
        (1 - M_{\text{GT}}) \odot A_{\text{origin}}, 
        & \text{if } R < 0.5, \\[3pt]
        A_{\text{origin}}, 
        & \text{otherwise},
    \end{cases}
\end{equation}
\noindent where \(\odot\) denotes the Hadamard (element-wise) product, and a random variable $R \sim U(0, 1)$. To preserve features of ID object, we utilize the ground-truth OOD mask \( M_{\text{GT}} \).









\begin{figure*}[b]
    \centering
    \includegraphics[width=\textwidth]{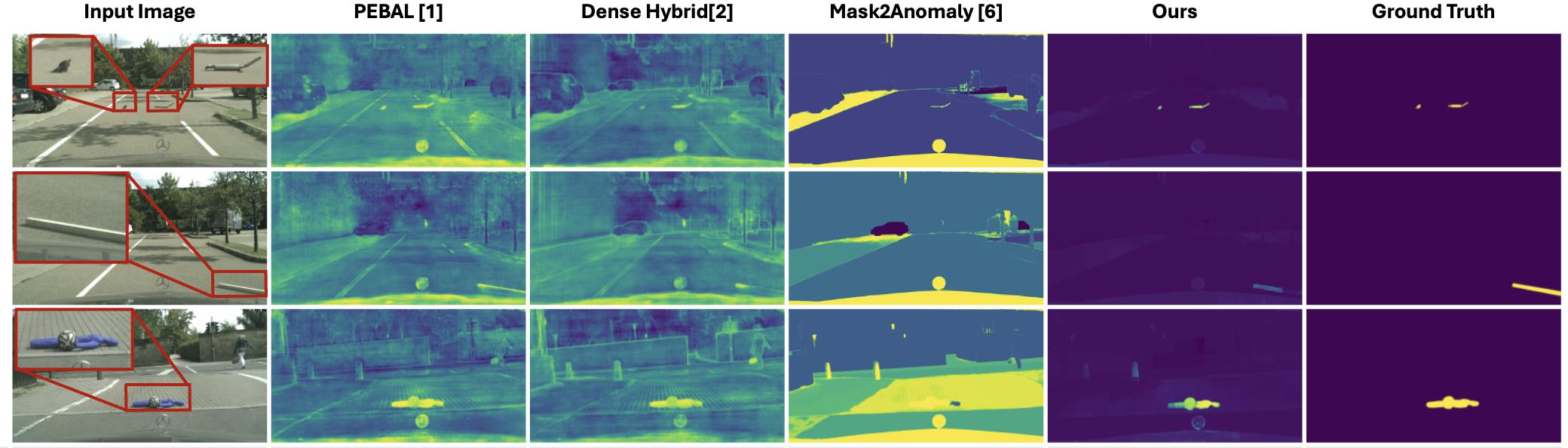} 
    \caption{Comparison of OOD segmentation visualization results. The input image (highlighting the OOD object in red box) and its corresponding segmentation outputs illustrate that our method not only provides more refined OOD predictions but also exhibits fewer false positives and false negatives than recent alternative methods.}
    \label{fig:quality_results}
\end{figure*}

\begin{table*}[t]
    \centering
    \renewcommand{\arraystretch}{1.2}
    \setlength{\tabcolsep}{4pt} 
    \resizebox{\textwidth}{!}{ 
    \begin{tabular}{l c ccccccccccccc}
        \toprule
        \multirow{2}{*}{Methods} 
        & \multirow{2}{*}{FLOPs} 
        & \multicolumn{2}{c}{FS Static} 
        & \multicolumn{2}{c}{FS Lost\&Found} 
        & \multicolumn{2}{c}{SMIYC-Anomaly} 
        & \multicolumn{2}{c}{SMIYC-Obstacle} 
        & \multicolumn{2}{c}{Road Anomaly} 
        & \multicolumn{2}{c}{Average} \\
        \cmidrule(lr){3-4} \cmidrule(lr){5-6} \cmidrule(lr){7-8} \cmidrule(lr){9-10} \cmidrule(lr){11-12} \cmidrule(lr){13-14}
        & 
        & AuPRC $\uparrow$ & FPR95 $\downarrow$ 
        & AuPRC $\uparrow$ & FPR95 $\downarrow$ 
        & AuPRC $\uparrow$ & FPR95 $\downarrow$ 
        & AuPRC $\uparrow$ & FPR95 $\downarrow$ 
        & AuPRC $\uparrow$ & FPR95 $\downarrow$ 
        & AuPRC $\uparrow$ & FPR95 $\downarrow$ \\
        \midrule

        \textbf{SynBoost~\cite{synboost}} 
        & 476G & 72.59 & 18.75 & 43.22 & 15.79 & 56.44 & 46.86 & 71.34 & 31.77 & 48.85 & 31.77 & 67.18 & 28.19 \\

        \textbf{DenseHybrid~\cite{densehybrid}} 
        & 202G & 80.23 & 5.95 & 47.06 & \underline{3.97} & 77.09 & 9.81 & 87.08 & 0.24 & 45.10 & 44.58 & 67.15 & 13.11 \\

        \textbf{PEBAL~\cite{PEBAL}} 
        & 1373G & 92.38 & 1.73 & 44.17 & 7.58 & 49.14 & 14.60 & 86.44 & 0.68 & 45.10 & 44.58 & 47.15 & 31.47 \\

        \textbf{Mask2Anomaly~\cite{Mask2Anomaly}} 
        & 258G & \underline{95.20} & \underline{0.82} & 46.04 & 4.36 & \underline{88.7} & 14.60 & \underline{93.3} & 0.20 & \underline{79.70} & \underline{13.45} & 80.59 & 6.68 \\

        \textbf{RPL+CoroCL~\cite{RPL}} 
        & 1771G & 92.46 & \textbf{0.85} & \textbf{70.61} & \textbf{2.52} & 88.55 & \underline{7.18} & \textbf{96.91} & \underline{0.09} & 71.61 & 17.74 & \underline{84.03} & \underline{5.68} \\

        \textbf{S2M~\cite{S2M}} 
        & 2259G & 85.13 & 3.05 & 59.08 & 67.35 & 91.92 & 1.04 & 91.73 & \textbf{0.02} & 73.55 & 42.76 & 80.28 & 22.84  \\

        \textbf{Ours} 
        & 421G & \textbf{99.03} & 1.14 & \underline{69.73} & 15.04 & \textbf{93.89} & \textbf{1.13} & 95.25 & 1.78 & \textbf{81.62} & \textbf{8.02} & \textbf{87.90} & \textbf{5.42}  \\

        \bottomrule
    \end{tabular}
    } 
    \caption{Comparison of AuPRC and FPR95 metrics across various datasets for pixel-level OOD segmentation}
    \label{tab:results01}
\end{table*}

\begin{table*}[t]
    \centering
    \renewcommand{\arraystretch}{1.2}
    \setlength{\tabcolsep}{4pt}

    \resizebox{\textwidth}{!}{
    \Huge
    \begin{tabular}{l c ccccccccccccccccccc}
        \toprule
        \multirow{2}{*}{Methods} 
        & \multirow{2}{*}{FLOPs}
        & \multicolumn{3}{c}{FS Static} 
        & \multicolumn{3}{c}{FS Lost\&Found} 
        & \multicolumn{3}{c}{SMIYC-Anomaly} 
        & \multicolumn{3}{c}{SMIYC-Obstacle} 
        & \multicolumn{3}{c}{Road Anomaly} 
        & \multicolumn{3}{c}{Average} \\
        \cmidrule(lr){3-5} \cmidrule(lr){6-8} \cmidrule(lr){9-11} \cmidrule(lr){12-14} \cmidrule(lr){15-17} \cmidrule(lr){18-20}
        & 
        & AuIoU $\uparrow$ & IoU $\uparrow$ & mean F1 $\uparrow$ 
        & AuIoU $\uparrow$ & IoU $\uparrow$ & mean F1 $\uparrow$ 
        & AuIoU $\uparrow$ & IoU $\uparrow$ & mean F1 $\uparrow$ 
        & AuIoU $\uparrow$ & IoU $\uparrow$ & mean F1 $\uparrow$ 
        & AuIoU $\uparrow$ & IoU $\uparrow$ & mean F1 $\uparrow$ 
        & AuIoU $\uparrow$ & IoU $\uparrow$ & mean F1 $\uparrow$ \\
        \midrule

        \textbf{Synboost~\cite{synboost}} 
        & 476G & 22.85 & 41.94 & 31.34 
        & 11.45 & 25.14 & 15.37 
        & 23.54 & 40.24 & 31.85 
        & 25.15 & 10.93 & 15.37 
        & 19.50 & 27.22 & 29.33 
        & 20.50 & 29.09 & 24.65 \\
        
        \textbf{DenseHybrid~\cite{densehybrid}} 
        & 202G & 22.80 & 41.93 & 31.30 
        & 10.93 & 25.14 & 15.37 
        & 30.32 & 45.10 & 41.31 
        & 23.54 & 40.25 & 31.85 
        & 19.50 & 27.22 & 29.33 
        & 21.42 & 35.93 & 29.83 \\

        \textbf{PEBAL~\cite{PEBAL}} 
        & 1373G & 23.65 & 69.00 & 29.66 
        & 7.03 & 28.21 & 9.70 
        & 21.50 & 44.47 & 36.35 
        & 4.12 & 27.25 & 6.27 
        & 15.51 & 33.80 & 23.87 
        & 14.36 & 40.55 & 21.17 \\

        \textbf{Mask2Anomaly~\cite{Mask2Anomaly}} 
        & 258G & 6.00 & 11.60 & 10.57 
        & 0.68 & 1.57 & 1.29 
        & 44.08 & \underline{81.31} & 53.37 
        & 8.07 & 42.41 & 11.90 
        & 24.31 & 56.74 & 32.80 
        & 16.63 & 38.73 & 21.98 \\

        \textbf{RPL+CoroCL~\cite{RPL}} 
        & 1771G & 17.43 & 71.73 & 22.78 
        & 5.52 & 29.83 & 7.60 
        & 22.95 & 70.98 & 32.44 
        & 13.01 & 65.82 & 16.64 
        & 16.83 & 51.00 & 24.57 
        & 15.15 & 57.87 & 20.81 \\

        \textbf{S2M~\cite{S2M}} 
        & 2259G & \underline{72.54} & \underline{81.98} & \underline{76.57} 
        & \textbf{35.52} & \underline{34.93} & \textbf{41.52} 
        & \underline{58.55} & \textbf{85.06} & \underline{65.13} 
        & \textbf{64.80} & \underline{69.63} & \underline{68.73} 
        & \underline{49.09} & \underline{58.58} & \underline{55.73} 
        & \underline{56.10} & \underline{66.04} & \underline{61.54} \\

        \textbf{Ours} 
        & 421G & \textbf{83.92} & \textbf{93.99} & \textbf{87.95} 
        & \underline{18.85} & \textbf{58.33} & \underline{23.48} 
        & \textbf{58.72} & 70.70 & \textbf{66.91} 
        & \underline{63.77} & \textbf{83.93} & \textbf{72.38} 
        & \textbf{56.74} & \textbf{65.34} & \textbf{66.66} 
        & \textbf{56.40} & \textbf{74.46} & \textbf{63.48} \\
        \bottomrule
    \end{tabular}
    } 
    \caption{Comparison of AuIoU, IoU, and Mean F1 scores across various datasets for object-level OOD segmentation}
    \label{tab:results02}
\end{table*}

\section{EXPERIMENTS}

\begin{table*}[htbp]
    \centering
    \caption{Ablation Study}
    \label{tab:combined_layout}
    \begin{minipage}[t]{0.48\linewidth}
        \centering
        \captionsetup{type=table,labelformat=empty}
        \caption*{(a) Ablation study results between variants of our model}
        \resizebox{\linewidth}{1cm}{%
            \begin{tabular}{ccc|ccccc}
                \toprule
                \textbf{\emph{A}} & \emph{B} & \emph{C} & \textbf{AuPRC$\uparrow$} & \textbf{FPR95$\downarrow$} & \textbf{AuIoU$\uparrow$} & \textbf{IoU$\uparrow$} & \textbf{Mean F1$\uparrow$} \\
                \midrule
                \checkmark &  &  & 92.15  & 2.60  & 78.20  & 86.27  & 82.58  \\
                \checkmark & \checkmark &  & \underline{98.95}  & \underline{1.02}  & 82.19  & \textbf{89.91}  & \underline{86.96} \\
                \checkmark & \checkmark & \checkmark & \textbf{99.28}  & \textbf{0.93}  & \textbf{94.53}  & \underline{87.63}  & \textbf{91.84} \\
                \bottomrule
            \end{tabular}
        }
    \end{minipage}
    \hfill
    \begin{minipage}[t]{0.48\linewidth}
        \centering
        \captionsetup{type=table,labelformat=empty}
        \caption*{(c) Effect of $\lambda$ in \emph{Semantically Augmented Attention} $A_{\text{SAA}}$}
        \resizebox{\linewidth}{1cm}{%
            \begin{tabular}{lccccc}
                \toprule
                \textbf{$\lambda$} & \textbf{AuPRC$\uparrow$} & \textbf{FPR95$\downarrow$} & \textbf{AuIoU$\uparrow$} & \textbf{IoU$\uparrow$} & \textbf{Mean F1$\uparrow$} \\
                \midrule
                1.0     & \textbf{98.82} & \underline{0.84} & 81.37 & 93.12 & 86.19 \\
                0.1     & \underline{99.42} & \textbf{0.72} & \underline{87.75} & \underline{95.41} & \underline{91.86} \\
                0.01    & 99.24 & 1.21 & \textbf{88.00} & \textbf{95.72} & \textbf{91.93} \\            
                \bottomrule
            \end{tabular}
        }
    \end{minipage}
    
    \vspace{1em} 
    
    \begin{minipage}[t]{0.48\linewidth}
        \centering
        \captionsetup{type=table,labelformat=empty}
        \caption*{(b) Effect of leveraging \emph{OOD Semantic Augmentation}}
        \resizebox{\linewidth}{1.0cm}{%
            \begin{tabular}{cc|ccccc}
                \toprule
                \textbf{First Layer} & \textbf{Last Layer} & \textbf{AuPRC$\uparrow$} & \textbf{FPR95$\downarrow$} & \textbf{AuIoU$\uparrow$} & \textbf{IoU$\uparrow$} & \textbf{Mean F1$\uparrow$} \\
                \midrule
                \checkmark &  & 99.03 & \underline{1.14} & 83.92 & 83.92 & 87.95  \\
                & \checkmark & \underline{99.21} & \textbf{0.93} & \underline{93.26} & \underline{84.65} & \underline{89.95}  \\
                \checkmark & \checkmark & \textbf{99.28} & \textbf{0.93} & \textbf{94.53} & \textbf{87.63} & \textbf{91.84} \\
                \bottomrule
            \end{tabular}
        }
    \end{minipage}
    \hfill
    \begin{minipage}[t]{0.48\linewidth}
        \centering
        \captionsetup{type=table,labelformat=empty}
        \caption*{(d) Effect of $\vert$Queries$\vert$ in \emph{Distance-Based OOD Prompt}}
        \resizebox{\linewidth}{1cm}{%
            \begin{tabular}{cccccc}
                \toprule
                \textbf{\# of Query} & \textbf{AuPRC$\uparrow$} & \textbf{FPR95$\downarrow$} & \textbf{AuIoU$\uparrow$} & \textbf{IoU$\uparrow$} & \textbf{Mean F1$\uparrow$} \\
                \midrule
                1  & 98.95 & \textbf{1.02}  & 82.19 & 89.91 & 86.96 \\
                25 & 98.51 & 1.47  & 82.20 & 93.19 & 87.10 \\
                50 & \textbf{99.03} & \underline{1.14}  & \underline{83.92} & \underline{93.99} & \underline{87.95} \\
                75 & \underline{99.02} & 1.58  & \textbf{85.01} & \textbf{95.19} & \textbf{89.20} \\
                \bottomrule
            \end{tabular}
        }
    \end{minipage}
    \label{tab:ablation}
\end{table*}

To rigorously evaluate our proposed method, we first conduct experiments on multiple publicly available OOD segmentation datasets, demonstrating that our model meets or exceeds the performance of state-of-the-art methods. We then provide a detailed analysis on the well-known Fishyscapes Lost\&Found~\cite{Mask2Anomaly} dataset, further substantiating our claims.

\subsection{Implementation and Evaluation Details}



Our architecture is built on a ViT~\cite{ViT} encoder, specifically using the CLIP ViT-B/16~\cite{clip} pretrained on LAION-400M. The pixel decoder of our model uses Mask2Former~\cite{mask2former} and TQDM~\cite{TQDM}. For generating OOD labels, we randomly sample 10,000 words from WordNet~\cite{WordNet}, apply a filtering step, and then group them into 50 clusters to be integrated into the training step. During training, we set the initial learning rate to \(1\times e^{-4}\), use a batch size of 4, and adopt the AdamW optimizer with a weight decay of 0.05. The model is trained for a total of 10,000 iterations. 

We use the Cityscapes~\cite{CityScapes} dataset, which includes pixel-level semantic segmentation annotations of complex urban road scenes, as the primary training set. Following Mask2Anomaly~\cite{Mask2Anomaly}, we treat the Cityscapes dataset as the reference domain of normal urban driving scenarios. To enable the model to detect OOD objects, we insert objects from the COCO dataset~\cite{coco} into Cityscapes images, allowing the model to learn new object types and thus improve OOD recognition.

For model evaluation, we employ a variety of benchmark datasets that include different domain gaps, ensuring robust OOD segmentation under diverse conditions. Segment Me If You Can (SMIYC)~\cite{SMITC} is used to assess unknown object detection in road scenes containing unexpected objects and images from varied environments. Road Anomaly~\cite{RoadAnomaly}, a web-sourced dataset comprising road images with possible anomalous objects, tests the model’s applicability in real-world conditions. Finally, Fishyscapes~\cite{Pinggera2016LostAF}, which includes both synthetic data and real driving scenes with small-scale anomalies, evaluates detection performance for smaller OOD objects. The code for the experiments will be found at our GitHub repository.





\subsection{Performance Evaluation}
We evaluate OOD segmentation performance at both pixel and object levels. Specifically, for pixel-level evaluations, we measure Area under the Precision-Recall Curve (AuPRC) and False Positive Rate at a true positive rate of 95\% (FPR95). To address the shortcomings of pixel-level evaluations--particularly their tendency to overlook small anomalies and overemphasize larger ones--we include additional object-level evaluations. In particular, we also report Intersection over Union (IoU), area under the IoU curve (AuIoU), and the mean F1 score.

Our approach demonstrates state-of-the-art (SOTA) performance across multiple benchmarks. As summarized in Table~\ref{tab:results01}, our method outperforms existing competitors on both AuPRC and FPR95 for pixel-level evaluations. In particular, the AuPRC score is on average 3.87\% higher than that of the second-best method, indicating that our model excels at detecting OOD regions.

On the object-level evaluations reported in Table~\ref{tab:results02}, our method again yields superior results compared to prior work. Notably, we observe gains of 8.42\% in IoU and 1.94\% in mean F1, demonstrating effective balancing of false positives (FP) and false negatives (FN). This means that our method not only accurately localizes OOD regions but also achieves robust detection of OOD objects overall.





Consequently, our method consistently outperforms competing approaches at both pixel-level and object-level evaluations, further demonstrating its ability to deliver highly reliable anomaly detection in OOD segmentation tasks.


\subsection{Qualitative Results}
In Figure~\ref{fig:quality_results}, we provide a comparison of visualizations of PEBAL~\cite{PEBAL}, DenseHybrid~\cite{densehybrid}, Mask2Anomaly~\cite{Mask2Anomaly}, and ours. Compared to alternative state-of-the-art methods, our method produces outputs that closely match the ground-truth annotations. In particular, it demonstrates superior pixel-level segmentation compared to prominent methods such as PEBAL~\cite{PEBAL} and DenseHybrid~\cite{densehybrid}. Moreover, in contrast to the recent Mask2Anomaly~\cite{Mask2Anomaly}, our model does not require a refinement step during inference, yet yields fewer false positives by focusing specifically on OOD objects.

\subsection{Ablation Study}
In Table~\ref{tab:ablation}, we compare variants of our model by removing each component on the OOD dataset called Fishyscapes as following Mask2Anomaly~\cite{Mask2Anomaly}, where we analyzed our model in detail to support our claim.   

\textbf{Effect of Each Components} We investigate how the three principal components proposed in Table~\ref{tab:ablation}(a)--namely \emph{A. Text-Driven OOD Segmentation}, \emph{B. Distance-Based OOD Prompts}, and \emph{C. OOD Semantic Augmentation}--contribute to our results. We bulit up \emph{Text-Driven OOD Segmentation} and then incrementally add each of the three components.
While using only the baseline improves object-level metrics, it results in slightly lower pixel-level evaluations compared to Mask2Anomaly. By contrast, applying both \emph{Distance-Based OOD Prompts} and \emph{OOD Semantic Augmentation} further boosts performance at both the pixel and object levels. This indicates that explicitly modeling diverse semantic distances for OOD data effectively strengthens the model’s capacity to generalize to novel OOD objects.

\textbf{Effect of Leveraging OOD Semantic Augmentation} 
We examine where to apply \emph{OOD Semantic Augmentation} by varying \emph{Semantically Augmented Attention} $A_{\text{SAA}}$ insertion layer within the model, as illustrated in Table~\ref{tab:ablation}(b). When $A_{\text{SAA}}$ is applied to high-level layers, we observe higher object-level evaluation (e.g., AuIoU), suggesting that introducing semantic perturbations at higher layers helps retain an object’s structural coherence while effectively modifying its semantics. Ultimately, combining $A_{\text{SAA}}$ at both low-level and high-level layers yields the best performance.

\textbf{Effect of $\lambda$ in \emph{Semantically Augmented Attention} $A_{\text{SAA}}$} 
We also evaluate the effect of different noise strengths $\lambda$ injected through $A_{\text{SAA}}$, as summarized in Table~\ref{tab:ablation}(c). Lower noise levels lead to better object-level evaluations but can degrade pixel-level evaluations, while stronger noise promotes richer semantic variation for pixel-level detection but risks diminishing structural information at the object level. A moderate noise intensity thus provides a balanced trade-off between these two aspects of OOD segmentation performance.

\textbf{Effect of the Number of \emph{Distance-Based OOD Prompt} Queries} 
Lastly, we investigate how the number of queries used in \emph{Distance-Based OOD Prompts} influences the results. Increasing the query count generally raises AuPRC and object-level evaluations, yet slightly inflates FPR95, implying a potential increase in false positives. These findings highlight the necessity of choosing an optimal number of queries to effectively balance detection performance and misclassification risks.

\section{CONCLUSIONS}




In this paper, we present a novel OOD segmentation method built upon a Vision-Language Model (VLM) to overcome the limitations of purely image-only approaches in autonomous driving scenarios. Our approach leverages three key components to effectively enhance generalization to a broad spectrum of OOD objects: \emph{A. Text-Driven OOD Segmentation}, \emph{B. Distance-Based OOD Prompts}, and \emph{C. OOD Semantic Augmentation}. These components enable the model to robustly distinguish between in-distribution (ID) and out-of-distribution (OOD) objects, thereby substantially improving OOD segmentation performance.

Our experimental results show significant performance gains over state-of-the-art methods across multiple metrics--including AuPRC, FPR95, IoU, AuIoU, and mean F1--highlighting the potential of a robust OOD segmentation solution for enhancing the safety of autonomous driving systems. In conclusion, our findings demonstrate that vision-language–based OOD segmentation can provide practical safety and reliability in real-world autonomous driving environments.

\section{ACKNOWLEDGMENTS}
This work was supported by the Institute of Information \& Communications Technology Planning \& Evaluation (IITP) grant funded by the Korea government (MSIT) (No. RS-2025-02219317, AI Star Fellowship at Kookmin University), (No. RS-2025-02263754, Human-Centric Embodied AI Agents with Autonomous Decision-Making), and (No. RS-2024-00397085, Leading Generative AI Human Resources Development), also supported by the National Research Foundation of Korea (NRF) grant funded by the Korea government (MSIT) (No. RS-2023-00212484, xAI for Motion Prediction in Complex, Real-World Driving Environment).



\bibliographystyle{IEEEtran.bst}
\bibliography{reference}

@inproceedings{Mask2Former,
  title={Masked-attention mask transformer for universal image segmentation},
  author={Cheng, Bowen and Misra, Ishan and Schwing, Alexander G and Kirillov, Alexander and Girdhar, Rohit},
  booktitle={Proceedings of the IEEE/CVF conference on computer vision and pattern recognition},
  pages={1290--1299},
  year={2022}
}

@article{Mask2Anomaly,
  title={Mask2anomaly Mask transformer for universal open-set segmentation},
  author={Rai, Shyam Nandan and Cermelli, Fabio and Caputo, Barbara and Masone, Carlo},
  journal={IEEE Transactions on Pattern Analysis and Machine Intelligence},
  year={2024},
  publisher={IEEE}
}

@article{CLIP,
      title={Learning Transferable Visual Models From Natural Language Supervision}, 
      author={Alec Radford and Jong Wook Kim and Chris Hallacy and Aditya Ramesh and Gabriel Goh and Sandhini Agarwal and Girish Sastry and Amanda Askell and Pamela Mishkin and Jack Clark and Gretchen Krueger and Ilya Sutskever},
      year={2021},
      eprint={2103.00020},
      archivePrefix={arXiv},
      primaryClass={cs.CV},
      url={https://arxiv.org/abs/2103.00020}, 
}

@inproceedings{DenseCLIP,
  title={Denseclip: Language-guided dense prediction with context-aware prompting},
  author={Rao, Yongming and Zhao, Wenliang and Chen, Guangyi and Tang, Yansong and Zhu, Zheng and Huang, Guan and Zhou, Jie and Lu, Jiwen},
  booktitle={Proceedings of the IEEE/CVF conference on computer vision and pattern recognition},
  pages={18082--18091},
  year={2022}
}

@inproceedings{TQDM,
  title={Textual Query-Driven Mask Transformer for Domain Generalized Segmentation},
  author={Pak, Byeonghyun and Woo, Byeongju and Kim, Sunghwan and Kim, Dae-hwan and Kim, Hoseong},
  booktitle={European Conference on Computer Vision},
  pages={37--54},
  year={2024},
  organization={Springer}
}

@misc{MTA-CLIP,
      title={MTA-CLIP: Language-Guided Semantic Segmentation with Mask-Text Alignment}, 
      author={Anurag Das and Xinting Hu and Li Jiang and Bernt Schiele},
      year={2024},
      eprint={2407.21654},
      archivePrefix={arXiv},
      primaryClass={cs.CV},
      url={https://arxiv.org/abs/2407.21654}, 
}

@misc{POC,
      title={Placing Objects in Context via Inpainting for Out-of-distribution Segmentation}, 
      author={Pau de Jorge and Riccardo Volpi and Puneet K. Dokania and Philip H. S. Torr and Gregory Rogez},
      year={2024},
      eprint={2402.16392},
      archivePrefix={arXiv},
      primaryClass={cs.CV},
      url={https://arxiv.org/abs/2402.16392}, 
}

@misc{StableDiffusion,
      title={High-Resolution Image Synthesis with Latent Diffusion Models}, 
      author={Robin Rombach and Andreas Blattmann and Dominik Lorenz and Patrick Esser and Björn Ommer},
      year={2022},
      eprint={2112.10752},
      archivePrefix={arXiv},
      primaryClass={cs.CV},
      url={https://arxiv.org/abs/2112.10752}, 
}

@misc{ResNet,
      title={Deep Residual Learning for Image Recognition}, 
      author={Kaiming He and Xiangyu Zhang and Shaoqing Ren and Jian Sun},
      year={2015},
      eprint={1512.03385},
      archivePrefix={arXiv},
      primaryClass={cs.CV},
      url={https://arxiv.org/abs/1512.03385}, 
}

@misc{DeepLab,
      title={DeepLab: Semantic Image Segmentation with Deep Convolutional Nets, Atrous Convolution, and Fully Connected CRFs}, 
      author={Liang-Chieh Chen and George Papandreou and Iasonas Kokkinos and Kevin Murphy and Alan L. Yuille},
      year={2017},
      eprint={1606.00915},
      archivePrefix={arXiv},
      primaryClass={cs.CV},
      url={https://arxiv.org/abs/1606.00915}, 
}

@misc{swin,
      title={Swin Transformer: Hierarchical Vision Transformer using Shifted Windows}, 
      author={Ze Liu and Yutong Lin and Yue Cao and Han Hu and Yixuan Wei and Zheng Zhang and Stephen Lin and Baining Guo},
      year={2021},
      eprint={2103.14030},
      archivePrefix={arXiv},
      primaryClass={cs.CV},
      url={https://arxiv.org/abs/2103.14030}, 
}

@misc{ViT,
      title={An Image is Worth 16x16 Words: Transformers for Image Recognition at Scale}, 
      author={Alexey Dosovitskiy and Lucas Beyer and Alexander Kolesnikov and Dirk Weissenborn and Xiaohua Zhai and Thomas Unterthiner and Mostafa Dehghani and Matthias Minderer and Georg Heigold and Sylvain Gelly and Jakob Uszkoreit and Neil Houlsby},
      year={2021},
      eprint={2010.11929},
      archivePrefix={arXiv},
      primaryClass={cs.CV},
      url={https://arxiv.org/abs/2010.11929}, 
}

@misc{CityScapes,
      title={A Review on Deep Learning Techniques Applied to Semantic Segmentation}, 
      author={Alberto Garcia-Garcia and Sergio Orts-Escolano and Sergiu Oprea and Victor Villena-Martinez and Jose Garcia-Rodriguez},
      year={2017},
      eprint={1704.06857},
      archivePrefix={arXiv},
      primaryClass={cs.CV},
      url={https://arxiv.org/abs/1704.06857}, 
}

@misc{negativelabelguidedood,
      title={Negative Label Guided OOD Detection with Pretrained Vision-Language Models}, 
      author={Xue Jiang and Feng Liu and Zhen Fang and Hong Chen and Tongliang Liu and Feng Zheng and Bo Han},
      year={2024},
      eprint={2403.20078},
      archivePrefix={arXiv},
      primaryClass={cs.CV},
      url={https://arxiv.org/abs/2403.20078}, 
}

@inproceedings{wordnet,
    title = "{W}ord{N}et A Lexical Database for {E}nglish",
    author = "Miller, George A.",
    booktitle = "Speech and Natural Language: Proceedings of a Workshop Held at Harriman, New York, {F}ebruary 23-26, 1992",
    year = "1992",
    url = "https://aclanthology.org/H92-1116/"
}

@misc{SAM,
      title={Segment Anything}, 
      author={Alexander Kirillov and Eric Mintun and Nikhila Ravi and Hanzi Mao and Chloe Rolland and Laura Gustafson and Tete Xiao and Spencer Whitehead and Alexander C. Berg and Wan-Yen Lo and Piotr Dollár and Ross Girshick},
      year={2023},
      eprint={2304.02643},
      archivePrefix={arXiv},
      primaryClass={cs.CV},
      url={https://arxiv.org/abs/2304.02643}, 
}

@misc{ghosh2024exploringfrontiervisionlanguagemodels,
      title={Exploring the Frontier of Vision-Language Models: A Survey of Current Methodologies and Future Directions}, 
      author={Akash Ghosh and Arkadeep Acharya and Sriparna Saha and Vinija Jain and Aman Chadha},
      year={2024},
      eprint={2404.07214},
      archivePrefix={arXiv},
      primaryClass={cs.CV},
      url={https://arxiv.org/abs/2404.07214}, 
}

@misc{zhang2024visionlanguagemodelsvisiontasks,
      title={Vision-Language Models for Vision Tasks: A Survey}, 
      author={Jingyi Zhang and Jiaxing Huang and Sheng Jin and Shijian Lu},
      year={2024},
      eprint={2304.00685},
      archivePrefix={arXiv},
      primaryClass={cs.CV},
      url={https://arxiv.org/abs/2304.00685}, 
}

@misc{hendrycks2018baselinedetectingmisclassifiedoutofdistribution,
      title={A Baseline for Detecting Misclassified and Out-of-Distribution Examples in Neural Networks}, 
      author={Dan Hendrycks and Kevin Gimpel},
      year={2018},
      eprint={1610.02136},
      archivePrefix={arXiv},
      primaryClass={cs.NE},
      url={https://arxiv.org/abs/1610.02136}, 
}

@misc{robotood,
      title={Uncertainty Estimation and Out-of-Distribution Detection for LiDAR Scene Semantic Segmentation}, 
      author={Hanieh Shojaei and Qianqian Zou and Max Mehltretter},
      year={2024},
      eprint={2410.08687},
      archivePrefix={arXiv},
      primaryClass={cs.LG},
      url={https://arxiv.org/abs/2410.08687}, 
}

@misc{li2020outofdistributiondetectionskinlesion,
      title={Out-of-Distribution Detection for Skin Lesion Images with Deep Isolation Forest}, 
      author={Xuan Li and Yuchen Lu and Christian Desrosiers and Xue Liu},
      year={2020},
      eprint={2003.09365},
      archivePrefix={arXiv},
      primaryClass={cs.CV},
      url={https://arxiv.org/abs/2003.09365}, 
}

@misc{PEBAL,
      title={Pixel-wise Energy-biased Abstention Learning for Anomaly Segmentation on Complex Urban Driving Scenes}, 
      author={Yu Tian and Yuyuan Liu and Guansong Pang and Fengbei Liu and Yuanhong Chen and Gustavo Carneiro},
      year={2022},
      eprint={2111.12264},
      archivePrefix={arXiv},
      primaryClass={cs.CV},
      url={https://arxiv.org/abs/2111.12264}, 
}

@misc{ALIGN,
      title={Scaling Up Visual and Vision-Language Representation Learning With Noisy Text Supervision}, 
      author={Chao Jia and Yinfei Yang and Ye Xia and Yi-Ting Chen and Zarana Parekh and Hieu Pham and Quoc V. Le and Yunhsuan Sung and Zhen Li and Tom Duerig},
      year={2021},
      eprint={2102.05918},
      archivePrefix={arXiv},
      primaryClass={cs.CV},
      url={https://arxiv.org/abs/2102.05918}, 
}

@misc{coco,
      title={Microsoft COCO: Common Objects in Context}, 
      author={Tsung-Yi Lin and Michael Maire and Serge Belongie and Lubomir Bourdev and Ross Girshick and James Hays and Pietro Perona and Deva Ramanan and C. Lawrence Zitnick and Piotr Dollár},
      year={2015},
      eprint={1405.0312},
      archivePrefix={arXiv},
      primaryClass={cs.CV},
      url={https://arxiv.org/abs/1405.0312}, 
}

@misc{densehybrid,
      title={DenseHybrid: Hybrid Anomaly Detection for Dense Open-set Recognition}, 
      author={Matej Grcić and Petra Bevandić and Siniša Šegvić},
      year={2022},
      eprint={2207.02606},
      archivePrefix={arXiv},
      primaryClass={cs.CV},
      url={https://arxiv.org/abs/2207.02606}, 
}

@misc{synboost,
      title={Pixel-wise Anomaly Detection in Complex Driving Scenes}, 
      author={Giancarlo Di Biase and Hermann Blum and Roland Siegwart and Cesar Cadena},
      year={2021},
      eprint={2103.05445},
      archivePrefix={arXiv},
      primaryClass={cs.CV},
      url={https://arxiv.org/abs/2103.05445}, 
}

@misc{S2M,
      title={Segment Every Out-of-Distribution Object}, 
      author={Wenjie Zhao and Jia Li and Xin Dong and Yu Xiang and Yunhui Guo},
      year={2024},
      eprint={2311.16516},
      archivePrefix={arXiv},
      primaryClass={cs.CV},
      url={https://arxiv.org/abs/2311.16516}, 
}

@misc{RPL,
      title={Residual Pattern Learning for Pixel-wise Out-of-Distribution Detection in Semantic Segmentation}, 
      author={Yuyuan Liu and Choubo Ding and Yu Tian and Guansong Pang and Vasileios Belagiannis and Ian Reid and Gustavo Carneiro},
      year={2023},
      eprint={2211.14512},
      archivePrefix={arXiv},
      primaryClass={cs.CV},
      url={https://arxiv.org/abs/2211.14512}, 
}

@article{coop,
   title={Learning to Prompt for Vision-Language Models},
   volume={130},
   ISSN={1573-1405},
   url={http://dx.doi.org/10.1007/s11263-022-01653-1},
   DOI={10.1007/s11263-022-01653-1},
   number={9},
   journal={International Journal of Computer Vision},
   publisher={Springer Science and Business Media LLC},
   author={Zhou, Kaiyang and Yang, Jingkang and Loy, Chen Change and Liu, Ziwei},
   year={2022},
   month=jul, pages={2337–2348} }

@misc{cocoop,
      title={Conditional Prompt Learning for Vision-Language Models}, 
      author={Kaiyang Zhou and Jingkang Yang and Chen Change Loy and Ziwei Liu},
      year={2022},
      eprint={2203.05557},
      archivePrefix={arXiv},
      primaryClass={cs.CV},
      url={https://arxiv.org/abs/2203.05557}, 
}

@misc{maskclip,
      title={Extract Free Dense Labels from CLIP}, 
      author={Chong Zhou and Chen Change Loy and Bo Dai},
      year={2022},
      eprint={2112.01071},
      archivePrefix={arXiv},
      primaryClass={cs.CV},
      url={https://arxiv.org/abs/2112.01071}, 
}

@misc{SAM-CLIP,
      title={SAM-CLIP: Merging Vision Foundation Models towards Semantic and Spatial Understanding}, 
      author={Haoxiang Wang and Pavan Kumar Anasosalu Vasu and Fartash Faghri and Raviteja Vemulapalli and Mehrdad Farajtabar and Sachin Mehta and Mohammad Rastegari and Oncel Tuzel and Hadi Pouransari},
      year={2024},
      eprint={2310.15308},
      archivePrefix={arXiv},
      primaryClass={cs.CV},
      url={https://arxiv.org/abs/2310.15308}, 
}

@misc{FC-CLIP,
      title={Convolutions Die Hard: Open-Vocabulary Segmentation with Single Frozen Convolutional CLIP}, 
      author={Qihang Yu and Ju He and Xueqing Deng and Xiaohui Shen and Liang-Chieh Chen},
      year={2023},
      eprint={2308.02487},
      archivePrefix={arXiv},
      primaryClass={cs.CV},
      url={https://arxiv.org/abs/2308.02487}, 
}

@misc{VLTSeg,
      title={Strong but simple: A Baseline for Domain Generalized Dense Perception by CLIP-based Transfer Learning}, 
      author={Christoph Hümmer and Manuel Schwonberg and Liangwei Zhou and Hu Cao and Alois Knoll and Hanno Gottschalk},
      year={2024},
      eprint={2312.02021},
      archivePrefix={arXiv},
      primaryClass={cs.CV},
      url={https://arxiv.org/abs/2312.02021}, 
}

@misc{CLIPSeg,
      title={Image Segmentation Using Text and Image Prompts}, 
      author={Timo Lüddecke and Alexander S. Ecker},
      year={2022},
      eprint={2112.10003},
      archivePrefix={arXiv},
      primaryClass={cs.CV},
      url={https://arxiv.org/abs/2112.10003}, 
}

@misc{SSPrompt,
      title={Learning to Prompt Segment Anything Models}, 
      author={Jiaxing Huang and Kai Jiang and Jingyi Zhang and Han Qiu and Lewei Lu and Shijian Lu and Eric Xing},
      year={2024},
      eprint={2401.04651},
      archivePrefix={arXiv},
      primaryClass={cs.CV},
      url={https://arxiv.org/abs/2401.04651}, 
}

@article{Pinggera2016LostAF,
  title={Lost and Found: detecting small road hazards for self-driving vehicles},
  author={Peter Pinggera and Sebastian Ramos and Stefan K. Gehrig and Uwe Franke and Carsten Rother and Rudolf Mester},
  journal={2016 IEEE/RSJ International Conference on Intelligent Robots and Systems (IROS)},
  year={2016},
  pages={1099-1106},
  url={https://api.semanticscholar.org/CorpusID:1236166}
}

@misc{RoadAnomaly,
      title={Detecting the Unexpected via Image Resynthesis}, 
      author={Krzysztof Lis and Krishna Nakka and Pascal Fua and Mathieu Salzmann},
      year={2019},
      eprint={1904.07595},
      archivePrefix={arXiv},
      primaryClass={cs.CV},
      url={https://arxiv.org/abs/1904.07595}, 
}

@misc{SMITC,
      title={SegmentMeIfYouCan: A Benchmark for Anomaly Segmentation}, 
      author={Robin Chan and Krzysztof Lis and Svenja Uhlemeyer and Hermann Blum and Sina Honari and Roland Siegwart and Pascal Fua and Mathieu Salzmann and Matthias Rottmann},
      year={2021},
      eprint={2104.14812},
      archivePrefix={arXiv},
      primaryClass={cs.CV},
      url={https://arxiv.org/abs/2104.14812}, 
}

@misc{RbA,
      title={RbA Segmenting Unknown Regions Rejected by All}, 
      author={Nazir Nayal and Mısra Yavuz and João F. Henriques and Fatma Güney},
      year={2023},
      eprint={2211.14293},
      archivePrefix={arXiv},
      primaryClass={cs.CV},
      url={https://arxiv.org/abs/2211.14293}, 
}

@misc{haldimann2019iimaginederrordetection,
      title={This is not what I imagined: Error Detection for Semantic Segmentation through Visual Dissimilarity}, 
      author={David Haldimann and Hermann Blum and Roland Siegwart and Cesar Cadena},
      year={2019},
      eprint={1909.00676},
      archivePrefix={arXiv},
      primaryClass={cs.CV},
      url={https://arxiv.org/abs/1909.00676}, 
}

@misc{xia2020synthesizecomparedetectingfailures,
      title={Synthesize then Compare: Detecting Failures and Anomalies for Semantic Segmentation}, 
      author={Yingda Xia and Yi Zhang and Fengze Liu and Wei Shen and Alan Yuille},
      year={2020},
      eprint={2003.08440},
      archivePrefix={arXiv},
      primaryClass={cs.CV},
      url={https://arxiv.org/abs/2003.08440}, 
}

@misc{lee2018trainingconfidencecalibratedclassifiersdetecting,
      title={Training Confidence-calibrated Classifiers for Detecting Out-of-Distribution Samples}, 
      author={Kimin Lee and Honglak Lee and Kibok Lee and Jinwoo Shin},
      year={2018},
      eprint={1711.09325},
      archivePrefix={arXiv},
      primaryClass={stat.ML},
      url={https://arxiv.org/abs/1711.09325}, 
}

@misc{liang2020enhancingreliabilityoutofdistributionimage,
      title={Enhancing The Reliability of Out-of-distribution Image Detection in Neural Networks}, 
      author={Shiyu Liang and Yixuan Li and R. Srikant},
      year={2020},
      eprint={1706.02690},
      archivePrefix={arXiv},
      primaryClass={cs.LG},
      url={https://arxiv.org/abs/1706.02690}, 
}

@misc{tian2021weaklysupervisedvideoanomalydetection,
      title={Weakly-supervised Video Anomaly Detection with Robust Temporal Feature Magnitude Learning}, 
      author={Yu Tian and Guansong Pang and Yuanhong Chen and Rajvinder Singh and Johan W. Verjans and Gustavo Carneiro},
      year={2021},
      eprint={2101.10030},
      archivePrefix={arXiv},
      primaryClass={cs.CV},
      url={https://arxiv.org/abs/2101.10030}, 
}

@article{ha2024leveraging,
  title={Leveraging Inductive Bias in ViT for Medical Image Diagnosis},
  author={Ha, Jungmin and Yoon, Euihyun and Kim, Sungsik and Kim, Jinkyu and Lee, Jaekoo},
  year={2024}
}

@inproceedings{lee2025controllable,
  title={Controllable 3D Object Generation with Single Image Prompt},
  author={Lee, Jaeseok and Lee, Jaekoo},
  booktitle={International Conference on Pattern Recognition},
  pages={222--238},
  year={2025},
  organization={Springer}
}

@inproceedings{kwon2023mobile,
  title={Mobile accelerator exploiting sparsity of multi-heads, lines, and blocks in transformers in computer vision},
  author={Kwon, Eunji and Song, Haena and Park, Jihye and Kang, Seokhyeong},
  booktitle={2023 Design, Automation \& Test in Europe Conference \& Exhibition (DATE)},
  pages={1--6},
  year={2023},
  organization={IEEE}
}

@article{kim2024instance,
  title={Instance-Dependent Multi-Label Noise Generation for Multi-Label Remote Sensing Image Classification},
  author={Kim, Youngwook and Kim, Sehwan and Ro, Youngmin and Lee, Jungwoo},
  journal={IEEE Journal of Selected Topics in Applied Earth Observations and Remote Sensing},
  year={2024},
  publisher={IEEE}
}

@inproceedings{kim2023bridging,
  title={Bridging the gap between model explanations in partially annotated multi-label classification},
  author={Kim, Youngwook and Kim, Jae Myung and Jeong, Jieun and Schmid, Cordelia and Akata, Zeynep and Lee, Jungwoo},
  booktitle={Proceedings of the IEEE/CVF Conference on Computer Vision and Pattern Recognition},
  pages={3408--3417},
  year={2023}
}

\end{document}